\documentclass[journal]{IEEEtran}
\usepackage{cite}
\usepackage{tabularx} % Load the package in your preamble

\ifCLASSINFOpdf
\else
\fi
\usepackage{amsmath}
\usepackage{algorithmic}

\usepackage{listings}
\usepackage{color}

\definecolor{dkgreen}{rgb}{0,0.6,0}
\definecolor{gray}{rgb}{0.5,0.5,0.5}
\definecolor{mauve}{rgb}{0.58,0,0.82}

\usepackage{algorithm}

\usepackage{tikz}
\usetikzlibrary{matrix, positioning}
\begin{document}
%
% paper title
% Titles are generally capitalized except for words such as a, an, and, as,
% at, but, by, for, in, nor, of, on, or, the, to and up, which are usually
% not capitalized unless they are the first or last word of the title.
% Linebreaks \\ can be used within to get better formatting as desired.
% Do not put math or special symbols in the title.
\title{Optimizing Sparse Convolution on GPUs with CUDA for 3D Point Cloud Processing in Embedded Systems}
%
%
% author names and IEEE memberships
% note positions of commas and nonbreaking spaces ( ~ ) LaTeX will not break
% a structure at a ~ so this keeps an author's name from being broken across
% two lines.
% use \thanks{} to gain access to the first footnote area
% a separate \thanks must be used for each paragraph as LaTeX2e's \thanks
% was not built to handle multiple paragraphs
%

\author{\IEEEauthorblockN{1\textsuperscript{st} Chester Luo} \\
\IEEEauthorblockA{
\textit{Brion Technologies, an ASML Company}\\
518000, Shenzhen, China \\
chester.luo02@gmail.com} \\
\and
\IEEEauthorblockN{2\textsuperscript{nd} Kevin Lai} \\
\IEEEauthorblockA{
\textit{School of Mechanical Engineering and Automation, Northeastern University}\\
110000, Shenyang, China \\
manhoi.laai@outlook.com} \\
}

\maketitle

% As a general rule, do not put math, special symbols or citations
% in the abstract or keywords.
\begin{abstract}
The increasing utilisation of LiDAR and 3D sensors has made the analysis of 3D point clouds indispensable in various applications, including object detection and segmentation. In contrast to photos, point clouds possess distinctive computing issues due to their sparse nature and absence of a regular grid. Traditional Convolutional Neural Networks (CNNs) that are designed to perform well on dense data are not suitable for this particular task. The utilisation of sparse neural networks specifically engineered for the efficient handling of sparse data is imperative in this context. Nevertheless, the utilisation of neural networks on Graphics Processing Units (GPUs), particularly with the CUDA framework, presents challenges due to the irregular nature of point cloud data and the requirement for optimised memory access patterns. The present study focuses on enhancing the efficiency of sparse convolution operators for 3D point clouds on GPUs through the utilisation of CUDA technology. This paper presents a novel approach that combines the theoretical benefits of sparse neural networks with efficient GPU-based implementations. By doing so, it provides valuable insights and techniques for effectively using 3D point cloud analysis, hence enhancing the capabilities of object detection and segmentation in various fields.
\end{abstract}

% Note that keywords are not normally used for peerreview papers.

% For peer review papers, you can put extra information on the cover
% page as needed:
% \ifCLASSOPTIONpeerreview
% \begin{center} \bfseries EDICS Category: 3-BBND \end{center}
% \fi
%
% For peerreview papers, this IEEEtran command inserts a page break and
% creates the second title. It will be ignored for other modes.
\IEEEpeerreviewmaketitle

\section{Introduction}
% The very first letter is a 2 line initial drop letter followed
% by the rest of the first word in caps.
% 
% form to use if the first word consists of a single letter:
% \IEEEPARstart{A}{demo} file is ....
% 
% form to use if you need the single drop letter followed by
% normal text (unknown if ever used by the IEEE):
% \IEEEPARstart{A}{}demo file is ....
% 
% Some journals put the first two words in caps:
% \IEEEPARstart{T}{his demo} file is ....
% 
% Here we have the typical use of a "T" for an initial drop letter
% and "HIS" in caps to complete the first word.
\IEEEPARstart{I}{n} recent years, there has been a significant increase in the utilization of deep learning methods, particularly convolutional neural networks (CNNs) \cite{articlecnn}, which have emerged as the dominant approach in various domains that involve structured grid data, such as picture analysis and processing. Nevertheless, the exponential growth in the utilization of LiDAR and 3D sensors across many domains has resulted in an increased need for the analysis of 3D point clouds. The utilization of 3D point clouds is crucial in various applications, including object recognition and segmentation, as they offer a spatial depiction of things within a three-dimensional environment. In contrast to photos, point clouds exhibit sparsity and lack a regular grid, hence posing distinct processing and computational issues.

The examination of 3D point cloud data has initiated a novel period of progress in the field of computer vision, enabling a wide range of applications that use its abundant spatial information. One of the most popular application of point cloud data is automated driving \cite{fernandes2021point} and deep learning can play role in various task on point cloud such as 3D shape classification, 3D object detection and tracking, 3D point cloud segmentation and 3D point cloud registration \cite{9127813}. Qi et al. introduced a significant contribution in the form of "PointNet" \cite{qi2017pointnet}, which marked a notable departure from existing approaches. This study presented a deep learning framework that was tailored to handle point sets, with a particular emphasis on addressing challenges related to 3D classification and segmentation tasks. In addition, Wang et al. introduced a real-time semantic segmentation method named "PointSeg" \cite{wang2018pointseg} that focuses on utilizing 3D LiDAR point clouds for this purpose. The aforementioned advancements highlight the extensive capabilities of 3D point cloud analysis in diverse practical situations, ranging from autonomous navigation to virtual reality. 

Point clouds, distinct from traditional image data due to their sparsity, present challenges in applying standard CNN architectures, which are better suited for dense data. This sparsity, a result of the data collection methodology and inherent information gaps in 3D space, necessitates the adoption of sparse neural networks. These networks are tailored to efficiently process sparse data while reducing computational demands. Sparse Convolution Neural Network (SCNN) is regarded as solution to applied deep learning on sparse data like point cloud. However, implementing them, particularly on GPUs, is complex due to the irregular data structures and the need for optimized memory access patterns. 

To bridge the gap between the computational intensity of SCNNs and the limited processing capabilities of edge computing devices, current developments in autonomous driving and 3D hand scanning technologies underscore the necessity for real-time processing of 3D point cloud data. These applications often operate in scenarios where rapid data processing is critical for decision-making and user interaction, yet they are constrained by the available computational resources. This dichotomy highlights the importance of efficient, real-time 3D data processing solutions that can operate within the confines of edge computing devices. NVIDIA's Jetson platform, equipped with CUDA technology, emerges as a suitable candidate for such tasks, offering a balance between computational power and energy efficiency, making it ideal for handling the demands of real-time, sparse data processing in applications like autonomous vehicles and advanced scanning devices. Recently, NVIDIA's introduction of the TensorRT \cite{vanholder2016efficient} inference framework has shown promise in addressing these issues. TensorRT, which allows custom plugin development, enhances the feasibility of deploying neural networks, specifically those utilizing sparse convolution, on embedded systems.

This research investigates the challenges of implement sparse convolution efficiently utilising GPUs on Jetson Platform with CUDA, to improve the speed of performing inference on sparse convolution operators for 3D point clouds. The main goal is to establish a correlation between the potential advantages of sparse neural networks and their practical application in real-life scenarios, and we will provide CUDA solutions for three categories of sparse convolution operators. This paper is structured as follows: Section II will provide a comparison between sparse convolution and traditional convolution to highlight their distinct characteristics. Additionally, it will present the current research on this topic. Section III presents a highly efficient method for implementing sparse convolution on CUDA and section IV outlines our discussions and conclusions.
% displays the experimental findings of our technique, while Section V 

\section{Background and Motivation}

\subsection{Structure and Represention for Point Cloud Data}

Point cloud data, a fundamental representation in 3D data analysis, consists of data points within a three-dimensional coordinate system, typically generated by 3D scanning technologies. These points, representing the external surfaces of objects, are characterized by their density, distribution, and precision. Traditional point cloud representation primarily involves raw coordinates, sometimes enriched with attributes like color, normal, and reflectance. However, the unstructured nature of raw point clouds poses challenges in computer vision and graphic because they are inherently complex and irregular, consist of a massive number of points and contain noise and outliers. Voxelization, the conversion of continuous point cloud data into a discrete 3D grid, addresses these challenges. Each voxel in the grid represents the presence or characteristics of points within its boundaries, thereby regularizing the data for computational efficiency in storage, compression, and processing. \cite{XU2021103675}

\begin{figure}[ht]
\includegraphics[width=8cm]{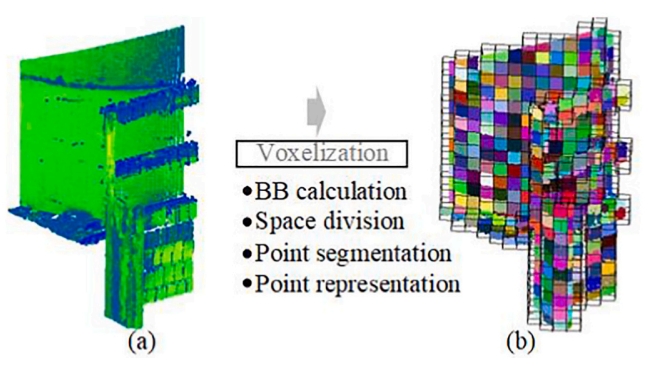}
\centering
\caption{Voxelization of 3D points: (a) Original TLS points; and (b) Voxelized point cloud. Source \cite{XU2021103675} The structured voxel grid offers a simplified and computationally efficient approach for 3D data analysis, suitable for applications in digital elevation modeling, urban planning, and 3D simulations.}
\end{figure}

When they need to be processed by neural network, for a point cloud and its voxelized data, the sparse tensor might encapsulate the actual 3D coordinates of points, alongside any associated attributes such as color, intensity, or normal. In typical sparse convolution implementations, data are stored in memory using specialized structures to efficiently represent and process the sparsity. The indices structure holds the coordinates of non-zero points in a sparse tensor, which corresponds to voxels or points in a point cloud. This structure allows the algorithm to iterate only over the meaningful data points, significantly reducing the computational load. For each batch in a dataset, the indices are stored with their batch ID, allowing for parallel processing of multiple data instances. The data structure for this work \cite{NEURIPS2019_57370345} \cite{article3dvoxel}

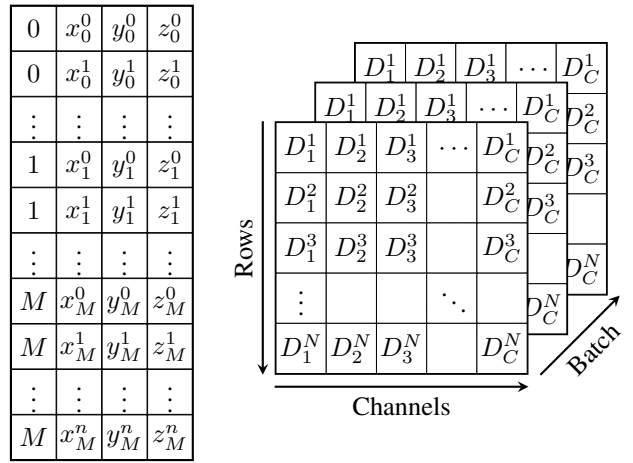
\begin{figure}[h]
    % \centering
    \begin{minipage}{0.15\linewidth}
    % \centering
\begin{tikzpicture}
    \matrix[matrix of math nodes,
    draw,thick, inner sep=0pt,
    nodes in empty cells,
            nodes={draw, very thin, minimum size=1.7em, anchor=center}] {
        0 & x_0^0 & y_0^0 & z_0^0 \\
        0 & x_0^1 & y_0^1 & z_0^1 \\
        \vdots & \vdots & \vdots & \vdots \\
        1 & x_1^0 & y_1^0 & z_1^0 \\
        1 & x_1^1 & y_1^1 & z_1^1 \\
        \vdots & \vdots & \vdots & \vdots \\
        M & x_M^0 & y_M^0 & z_M^0 \\
        M & x_M^1 & y_M^1 & z_M^1 \\
        \vdots & \vdots & \vdots & \vdots \\
        M & x_M^n & y_M^n & z_M^n \\
    };
\end{tikzpicture}

\end{minipage}
    \hspace{1.3cm} % <- Adjust this value for the desired spacing
    \begin{minipage}{0.15\linewidth}
\begin{tikzpicture}[auto matrix/.style={matrix of nodes,
  draw,thick,inner sep=0pt,
  nodes in empty cells,column sep=-0.2pt,row sep=-0.2pt,
  cells={nodes={minimum width=1.9em,minimum height=1.9em,
   draw,very thin,anchor=center,fill=white,
   execute at begin node={%
   $\vphantom{x_1^1}
    \pgfmathtruncatemacro{\itest}{sign(4-\the\pgfmatrixcurrentcolumn)*sign(4-\the\pgfmatrixcurrentrow)}
    \unless\ifnum\itest=0
    {#1}^{\myrowindex{\the\pgfmatrixcurrentrow}}_{\mycolindex{\the\pgfmatrixcurrentcolumn}}
    \fi
    \ifnum\the\pgfmatrixcurrentrow\the\pgfmatrixcurrentcolumn=14
    \cdots
    \fi
    \ifnum\the\pgfmatrixcurrentrow\the\pgfmatrixcurrentcolumn=41
    \vdots
    \fi
    \ifnum\the\pgfmatrixcurrentrow\the\pgfmatrixcurrentcolumn=44
    \ddots
    \fi
   $}
  }}}]
 \newcommand{\mycolindex}[1]{\ifnum#1=5 C\else #1\fi}
 \newcommand{\myrowindex}[1]{\ifnum#1=5 N\else #1\fi}
 \matrix[auto matrix=D,xshift=3em,yshift=3em](matz){
  & & & & \\
  & & & & \\
  & & & & \\
  & & & & \\
  & & & & \\
 };
 \matrix[auto matrix=D,xshift=1.5em,yshift=1.5em](maty){
  & & & & \\
  & & & & \\
  & & & & \\
  & & & & \\
  & & & & \\
 };
 \matrix[auto matrix=D](matx){
  & & & & \\
  & & & & \\
  & & & & \\
  & & & & \\
  & & & & \\
 };
 \draw[thick,-stealth] ([xshift=1ex]matx.south east) -- ([xshift=1ex]matz.south east)
  node[midway,below, rotate=45] {Batch};
 \draw[thick,-stealth] ([yshift=-1ex]matx.south west) -- 
  ([yshift=-1ex]matx.south east) node[midway,below] {Channels};
 \draw[thick,-stealth] ([xshift=-1ex]matx.north west)
   -- ([xshift=-1ex]matx.south west) node[midway,above,rotate=90] {Rows};
\end{tikzpicture}
    \end{minipage}
    
\caption{Illustration of the data structures for sparse tensor for voxel or point cloud. On the left is the indices structure, depicting the point coordinates (m, x, y, z) for each batch ranging from 0 to M+1, where m is the batch ID. On the right is the features structure, showcasing the organization of feature values across channels, rows, and batches.}
\end{figure}

\subsection{Dense Convolution v.s. Sparse Convolution}

In machine learning, 'sparse' and 'dense' matrices or tensors differ in their non-zero element distribution. Dense matrices/tensors are predominantly non-zero, implying rich information in almost every entry. In contrast, sparse matrices/tensors have a substantial proportion of zeros (or a default value), offering unique storage and computational challenges. Dense structures usually require resources proportional to their size for storage and processing. However, sparse structures benefit from specialized formats like Compressed Sparse Row (CSR)\cite{barrett1994templates} or ELLPACK Sparse Block (ESB)\cite{10.1145/2464996.2465013}, optimizing non-zero element storage and computation, thereby enhancing space efficiency.

\begin{figure}[!ht]
\centering
\begin{tikzpicture}[
    2d-arr/.style={matrix of nodes, row sep=-\pgflinewidth, column sep=-\pgflinewidth, nodes={draw}}
  ]

  \matrix (azx) [2d-arr] {
  |[fill=yellow!50]|3 & |[fill=yellow!50]|1 &|[fill=yellow!50]| 1 &|[fill=yellow!50]| 1 & 0 & |[fill=yellow!50]|2 & 0 \\
  |[fill=yellow!50]|2 & 0 &|[fill=yellow!50]| 1 &|[fill=yellow!50]| 1 &|[fill=yellow!50]| 1 & 0 & 0 \\
  |[fill=yellow!50]|1 & |[fill=yellow!50]|2 & 0 &|[fill=yellow!50]| 1 &|[fill=yellow!50]| 1 &|[fill=yellow!50]| 1 & |[fill=yellow!50]|1\\
  |[fill=yellow!50]|4 & |[fill=yellow!50]|3 & 0 &|[fill=yellow!50]| 1 & |[fill=yellow!50]|1 & 0 & 0\\
  |[fill=yellow!50]|3 & 0 & |[fill=yellow!50]|1 & |[fill=yellow!50]|1 & 0 & |[fill=yellow!50]|2 &|[fill=yellow!50]| 4\\
  0 &|[fill=yellow!50]| 1 & |[fill=yellow!50]|1 & 0 &|[fill=yellow!50]| 4 & 0 &|[fill=yellow!50]| 5\\
  |[fill=yellow!50]|1 &|[fill=yellow!50]| 1 & 0 & 0 &|[fill=yellow!50]| 5 & 0 & 0\\
  };

  \node[below=of azx-5-4] {2D dense matrix};

  \matrix (mtr) [2d-arr,  right=3em of azx] {
  0 & |[fill=yellow!50]|1 & 0 & |[fill=yellow!50]|1 & 0 & 0 & 0 \\
  0 & 0 & 0 & 0 & 0 & 0 & 0 \\
  0 & 0 & 0 & 0 & 0 & |[fill=yellow!50]|1 & |[fill=yellow!50]|1\\
  0 & 0 & 0 & 0 &|[fill=yellow!50]| 1 & 0 & 0\\
  0 & 0 & |[fill=yellow!50]|1 & 0 & 0 & 0 & 0\\
  0 & 0 & |[fill=yellow!50]|1 & 0 & 0 & 0 & 0\\
  |[fill=yellow!50]|1 & |[fill=yellow!50]|1 & 0 & 0 & 0 & 0 & 0\\
  };

  \node[below=of mtr-5-4] {2D sparse matrix};
\end{tikzpicture}

\caption{Illustrative comparison of a 2D dense matrix and a 2D sparse matrix. The dense matrix predominantly features non-zero elements, whereas the sparse matrix consists mainly of zero values with a few non-zero entries scattered throughout. Such distinctions highlight the storage and computational differences between the two matrix types.}
\label{fig_sim}
\end{figure}

In conventional convolutional procedures applied to dense data, a filter or kernel traverses the input data, often an image, resulting in the generation of an output feature map. The procedure is performing element-wise multiplication between the kernel and a specific local region of the input, and subsequently calculating the sum of the resulting products. In order to enhance the efficiency of this computation, a commonly utilized technique known as \texttt{im2col} is implemented. The \texttt{im2col} technique is employed to transform localized areas of the input data into column vectors, effectively transforming the convolution operation into a matrix multiplication, which is especially friendly to GPU by using GEMM \cite{nvidiaMatrixMultiplication}.

\begin{figure}[ht]
    \centering
    \includegraphics[width=0.9\linewidth]{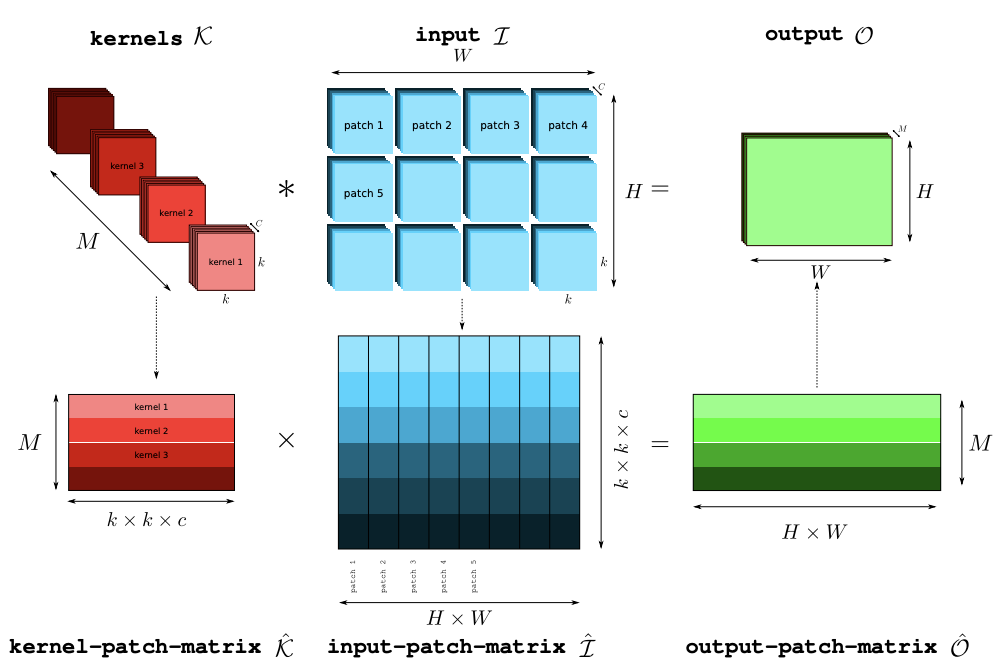}
    \caption{Illustration of the convolution process using the `im2col` approach, source \cite{vasudevan2017parallel}. Starting with a set of kernels $K$ and an input $I$, the method reshapes the input into overlapping patches. Correspondingly, kernels are reshaped into rows to form the kernel-patch-matrix $K'$. Matrix multiplication between $K'$ and the input-patch-matrix $I'$ yields the output-patch-matrix $O'$. The final output $O$ is derived by reshaping $O'$ to its intended dimensions.}
    \label{fig:convolution_im2col}
\end{figure}

Although conventional convolution is effective for dense data, it struggles with sparse datasets such as point clouds as it is not applicable to the irregular, sparse nature of point clouds. Consequently, these methods, including \texttt{im2col}, often process extensive empty spaces in sparse data, resulting in increased computational overhead and memory usage. To delve into the computational complexity, consider a typical convolution operation that might require \(O(n^2)\) operations for a dense \(n \times n\) matrix. For sparse data with only \(k\) non-zero elements, an optimal approach would ideally require close to \(O(k)\) operations. However, conventional convolution using techniques like \texttt{im2col} still approach a complexity of \(O(n^2)\), even though a vast majority of the processed elements are zeros. This means we're expending computational effort on a high proportion of elements that contribute no meaningful information to the output. Such inefficiencies underline the inadequacy of traditional convolution methods for sparse datasets and emphasize the need for specialized techniques tailored to the intricacies of sparse data.

To solve this inconvenience, Liu et al. introduced sparse convolutional neural networks (SCNN) in 2015 \cite{liu2015sparse}, an innovative approach leveraging a two-stage decomposition to reduce inter-channel and intra-channel redundancies in kernel weights. The process begins with an initial decomposition based on kernel weight reconstruction error, followed by a fine-tuning stage incorporating a sparsity constraint. This dual-phase approach effectively optimizes training error, kernel sparsity, and the number of convolutional bases. Further delving into the SCNN model, the authors elucidate that each sparse convolutional layer utilizes a select few kernels followed by sparse matrix multiplication. While this approach suggests inherent computational efficiency, the authors acknowledge potential overheads in sparse matrix computations. To mitigate this, they propose an optimized algorithm for sparse matrix multiplication, capitalizing on the fixed nature of sparse kernels post-training to avoid indirect and discontinuous memory access challenges. This CPU-focused strategy outperforms standard sparse matrix libraries, offering significant acceleration over traditional dense networks.

In 2017, a new sparse convolution method named 
Submanifold\cite{DBLP:journals/corr/GrahamM17} are introduced. This type of convolution operates exclusively on the spatial locations where input features are present, entirely bypassing the regions devoid of information. SubM convolution leverages the spatial sparsity to reduce computational requirements, ensuring that only relevant regions of the data are processed. This efficiency makes it particularly suitable for point cloud data where vast portions may lack relevant information. Based on this, Graham et al. presented a novel approach to 3D semantic segmentation by employing Submanifold Sparse Convolutional Networks \cite{graham20183d} in 2018, show the effective utilization of the distinctive characteristics of 3D data to achieve accurate.

\begin{figure}[!ht]
\centering
\begin{tikzpicture}[
    2d-arr/.style={matrix of nodes, row sep=-\pgflinewidth, column sep=-\pgflinewidth, nodes={draw}}
  ]

  \matrix (mtr) [2d-arr] {
  0 & 1 & 1 & |[fill=orange!30]| 1 & |[fill=orange!30]| 0 & |[fill=orange!30]| 0 & 0\\
  0 & 0 & 1 & |[fill=orange!30]| 1 & |[fill=orange!30]| 1 & |[fill=orange!30]| 0 & 0\\
  0 & 0 & 0 & |[fill=orange!30]| 1 & |[fill=orange!30]| 1 & |[fill=orange!30]| 1 & 0\\
  |[fill=red!30]|0 & |[fill=red!30]|0 & |[fill=red!30]|0 & 1 & 1 & 0 & 0\\
  |[fill=red!30]|0 & |[fill=red!30]|0 & |[fill=red!30]|1 & 1 & 0 & 0 & 0\\
  |[fill=red!30]|0 & |[fill=red!30]|1 & |[fill=red!30]|1 & 0 & 0 & 0 & 0\\
  1 & 1 & 0 & 0 & 0 & 0 & 0\\
  };

  \node[below=of mtr-5-4] {$\mathbf I$};

  \node[right=0.2em of mtr] (str) {$*$};

  \matrix (K) [2d-arr, right=0.2em of str, nodes={draw, fill=teal!30}] {
    1 & 0 & 1 \\
    0 & 1 & 0 \\
    1 & 0 & 1 \\
  };
  \node[below=of K-3-2] {$\mathbf K$};

  \node[right=0.2em of K] (eq) {$=$};

  \matrix (ret) [2d-arr, right=0.2em of eq] {
  0 & 4 & 3 & |[fill=blue!80!black!30]| 4 & 0\\
  0 & 0 & 4 & 3 & 0\\
  0 & 0 & 3 & 4 & 0\\
  0 & 3 & 3 & 0 & 0\\
  3 & 3 & 0 & 0 & 0\\
  };
  \node[below=of ret-4-3] {$\mathbf{I * K}$};

  \draw[dashed, teal] (mtr-1-6.north east) -- (K-1-1.north west);
  \draw[dashed, teal] (mtr-3-6.south east) -- (K-3-1.south west);

  \draw[dashed, blue!80!black] (K-1-3.north east) -- (ret-1-4.north west);
  \draw[dashed, blue!80!black] (K-3-3.south east) -- (ret-1-4.south west);

  \foreach \i in {1,2,3} {
      \foreach \j in {4,5,6} {
          \node[font=\tiny, scale=0.6, shift={(-1.2ex,-2ex)}] at (mtr-\i-\j) {$\times \pgfmathparse{int(mod(\i+\j,2))}\pgfmathresult$};
        }
    }

\end{tikzpicture}
\caption{Illustration of submanifold convolution on a 2D matrix. The input matrix \(I\) is convolved with the kernel \(K\). Only the central non-zero elements (highlighted in orange) undergo computation. Regions with a central zero (highlighted in red) are excluded from the convolution process.}
\label{fig_sim}
\end{figure}
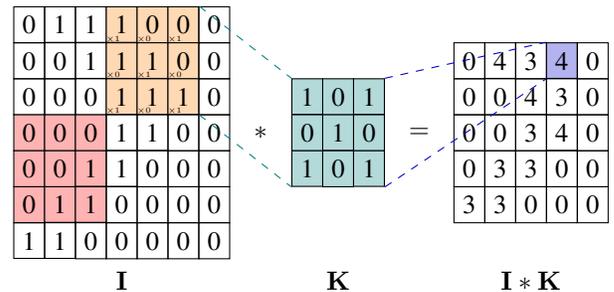

\subsection{Related Work}
In 2018, Elgammal et al. proposed a framework named uSCNN \cite{elgammal2023cuscnn}, an effective sparse CNN inference engine that takes use of the sparsity of models and activations, utilizing optimized sparse-sparse matrix convolution kernels with compressed operands. This work optimized the sparse CNN on image. In the realm for point cloud, Yan et al. introduces a general process to compute sparse convolution includes building rules table and set input values for GEMM process \cite{s18103337}. This work presents a comprehensive workflow detailing the entire process of computation within the realm of sparse data from LiDAR on GPU. Their work includes two stages: (1) Generate computation Rules (Algorithm 1). (2) Compute values based on rules. The algorithm 1 desribes the whole process and their code has been open-source on GitHub, names “SpConv” \cite{githubGitHubTraveller59spconv}. 

\begin{algorithm}
\caption{Rule Generation Process in \cite{s18103337}}
\begin{algorithmic}[1]
\REQUIRE Indice: Coordinates of active points, with dimensions of $n_{in} \times 3$; \\
Grid: A buffer with dimensions of $N \times D \times H \times W$; \\
$i_{in}$: The input index; \\
$i_{out}$: The output index;
\ENSURE Rule
\STATE Rule[:] $\leftarrow -1$
\STATE $n_{in} \leftarrow$ number of input points
\STATE $n_{kernel} \leftarrow$ volume of convolution kernel
\STATE $n_{out} \leftarrow$ number of output points
\FOR{$i_{in} = 0$ to $n_{in}$}
    \STATE $p_{in} \leftarrow$ Indice[$i_{in}$]
    \STATE $p_{out} \leftarrow$ getOutputCoord($p_{in}$)
    \FOR{p $\in p_{out}$}
        \STATE index $\leftarrow$ getSpatialIndex(p)
        \STATE offset $\leftarrow$ getOffset(p, $p_{in}$)
        \STATE Rule[offset, $i_{in}$, 0] $\leftarrow i_{in}$
        \STATE Rule[offset, $i_{in}$, 1] $\leftarrow$ index
    \ENDFOR
\ENDFOR
\STATE SpatialIndex $\leftarrow$ Rule[:, :, 1]
\STATE $n_{out} \leftarrow$ unique(SpatialIndex)
\FOR{$i_{out} = 0$ to $n_{out}$}
    \STATE Grid[SpatialIndex[$i_{out}$]] $\leftarrow i_{out}$
\ENDFOR
\FOR{$i_{in} = 0$ to $n_{in}$}
    \FOR{$j = 0$ to $n_{kernel}$}
        \STATE index $\leftarrow$ Rule[$j, i_{in}, 1$]
        \IF{index $> 0$}
            \STATE Rule[$j, i_{in}, 1$] $\leftarrow$ Grid[index]
        \ENDIF
    \ENDFOR
\ENDFOR
\label{algo:1}
\end{algorithmic}
\end{algorithm}

In 2022, Tang et al. introduced a new point cloud inference engine named “TorchSparse" \cite{tang2022torchsparse}, which has faster speed than "SpConv". This work introduces innovative matrix multiplication grouping strategies that target efficiency in computational tasks. The proposed fixed grouping strategy compensates for the excess FLOPs with enhanced regularity, while the adaptive grouping seeks an optimal balance point. Once the maps are devised, sparse convolution is employed, further enhancing the computation efficiency. Significantly, the utilization of matrix-vector multiplication on GPUs has been traditionally low. Addressing this, the work integrates a gather-matmul-scatter computation flow, ensuring a more streamlined and efficient process. Their work are also open-source on Github \cite{githubGitHubMithanlabtorchsparse}

\begin{figure}[!ht]
\centering
\includegraphics[width=0.5\textwidth]{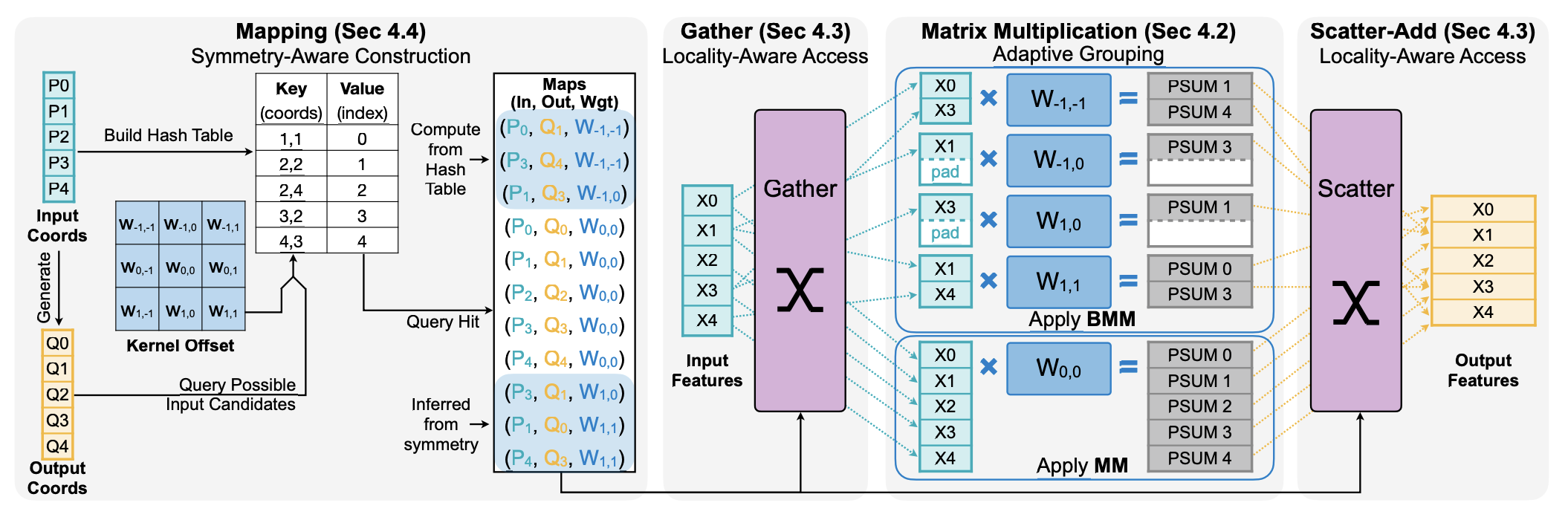}
\caption{TorchSparse concentrates on accelerating Sparse Convolution, delineated into four key stages: mapping, gather, matrix multiplication, and scatter-accumulate. Its objectives include augmenting the uniformity of sparse workloads and reducing memory consumption. These aims are realized through the use of adaptively batched matrix multiplication and the implementation of quantized, vectorized, locality-aware scatter/gather operations. Furthermore, kernel fusion in the mapping stage enhances memory efficiency.}
\label{fig_sim}
\end{figure}

However, both "SpConv" and “TorchSparse" are specific designed for PyTorch \cite{NEURIPS2019_9015}. The design innovations in Spconv and TorchSparse, tailored to PyTorch's unique features and data types, have significantly enhanced this environment's capabilities. Yet, their PyTorch-centric approach, beneficial in some aspects, also marks their fundamental limitations. These systems, designed specifically for PyTorch, lack flexibility for use with other platforms or frameworks. This issue is especially evident when deploying in varied computational environments or integrating with non-PyTorch systems. Our research aims to overcome these limitations by developing a version of sparse convolution that maintains the computational efficiency of Spconv and TorchSparse while also being compatible with a wider array of platforms and frameworks. This approach broadens their applicability and increases deployment flexibility.

\section{Our Approach}
This section will present the methodology employed for designing operators to maximize parallelism in CUDA, utilizing fundamental CUDA optimization strategies. Parallelism in CUDA is attained through the simultaneous execution of numerous threads, organized into blocks, across the multiple cores of a GPU. In addition to the aforementioned considerations, it is imperative to also prioritize the optimization of data load efficiency. Data locality refers to the practice of reusing data that is already fetched into the faster memory hierarchies, minimizing the need to access the slower global memory. In contrast, shared memory plays a crucial role in facilitating inter-thread communication and data exchange as it is accessible to all threads inside a block\cite{garland2008parallel}. By effectively utilizing both local and shared memory, it is possible to enhance data access patterns, decrease memory latency, and lessen dependence on the comparatively slower global memory. 

\subsection{Input and Output Structure}
The aforementioned publications discuss sparse tensor structures, which consist of two main data tensors: one that represents the indices of the points, and another that contains the related data for these points. This structure is developer-friendly and efficient in PyTorch when constructing and training networks. In order to incorporate this functionality into CUDA for embedded systems, it becomes necessary to modify these tensors to conform to conventional CUDA data types. The architecture of our operator utilises a collection of CUDA arrays consisting of integers (int) to represent the indices, along with another set of CUDA arrays for data (float). This approach effectively replaces the usual data tensors.

A critical issue in this context is maintaining the correspondence between input and output indices in typical sparse convolution (down-sampling) and efficiently using this mapping in inverse convolution (up-sampling). Our strategy emphasizes simplicity and efficiency. Instead of complicating the mapping process, we use the indices as direct input variables. An additional phase in the operator kernels then re-establishes their interconnection, ensuring efficient operation and optimal performance, we will discuss this in section IV.

To further improve the efficient of load input indices, we make a little change in input format to achieve coalesced access. In coalesced access, consecutive threads access consecutive memory locations, facilitating a linear and contiguous memory access pattern. This approach not only optimizes memory bandwidth utilization but also significantly bolsters kernel performance \cite{Kim2017EvaluationOT}.

\begin{figure}[h]
    
\begin{minipage}{0.5\textwidth}
\centering
    
\begin{tikzpicture}[
    2d-arr/.style={matrix of nodes,
        row sep=-\pgflinewidth,
        column sep=-\pgflinewidth, 
        nodes={draw}}
  ]

\matrix (mtr) [2d-arr,nodes={draw,
                      minimum width=1.9em,
                      minimum height=1.9em,
                      anchor=center,
                      inner sep=0pt,
                      outer sep=0pt,
                      font=\scriptsize},
               column sep=-\pgflinewidth,
               row sep=-\pgflinewidth] { 
 |[fill=blue!30]|$i_0$ & |[fill=blue!30]|$i_1$ & |[fill=blue!30]|$i_2$ & |[fill=blue!30]|$i_3$ & |[fill=blue!30]|$i_4$ &|[fill=blue!30]|$\cdots$ & |[fill=blue!30]|$i_{n-1}$ \\ 
  |[fill=yellow!30]|$x_0$ & |[fill=yellow!30]|$x_1$ & |[fill=yellow!30]|$x_2$ & |[fill=yellow!30]|$x_3$ & |[fill=yellow!30]|$x_4$ &|[fill=yellow!30]|$\cdots$ & |[fill=yellow!30]|$x_{n-1}$ \\ 
  |[fill=lime!30]|$y_0$ & |[fill=lime!30]|$y_1$ & |[fill=lime!30]|$y_2$ & |[fill=lime!30]|$y_3$ & |[fill=lime!30]|$y_4$ &|[fill=lime!30]|$\cdots$ & |[fill=lime!30]|$y_{n-1}$ \\ 
    |[fill=teal!30]|$z_0$ & |[fill=teal!30]|$z_1$ & |[fill=teal!30]|$z_2$ & |[fill=teal!30]|$z_3$ & |[fill=teal!30]|$z_4$ &|[fill=teal!30]|$\cdots$ & |[fill=teal!30]|$z_{n-1}$ \\ 
};

% y-axis
\draw[thick,->] ([xshift=-1em]mtr.north west) -- ([xshift=-1em]mtr.south west) node[left] {Access Direction};

\end{tikzpicture}
\end{minipage}
\caption{The memory layout of indices in CUDA. This configuration ensures an optimized memory access pattern, pivotal for efficient CUDA processing.}
\end{figure}
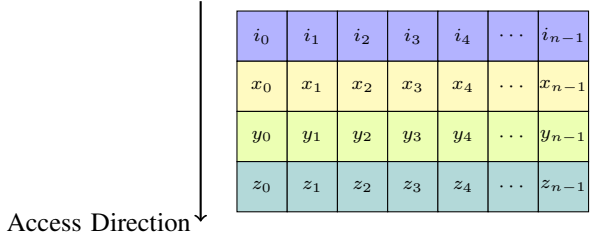

\subsection{Sampling Indices Mapping}

As we get a list of points coordinates with no order, it is essential to determine a rules to help us record these points and find a better representation in computing process. Given a point's position before convolution, denoted as $(x, y, z)$, our objective is to establish its relationship post-convolution. The following steps elucidate this process:

\subsubsection{Position Identification}

Determine if the point is at the start of a convolution cell using the divisibility condition:
\begin{equation}
\begin{aligned}
    x \mod s &= 0 , 
    y \mod s &= 0, 
    z \mod s &= 0,
\end{aligned}
\end{equation}
where $s$ is the stride size.

\subsubsection{Start Cell Determination}

If the point satisfies the above condition, it lies at the start of a cell. Otherwise, ascertain the start cell's coordinates as:
\begin{equation}
\begin{aligned}
    startCell_X &= \left\lfloor \frac{x}{s} \right\rfloor \times s, \\
    startCell_Y &= \left\lfloor \frac{y}{s} \right\rfloor \times s, \\
    startCell_Z &= \left\lfloor \frac{z}{s} \right\rfloor \times s.
\end{aligned}
\end{equation}

\subsubsection{Offsets Calculation}

Calculate the point's offset within its respective cell:
\begin{equation}
\begin{aligned}
    offset_X &= x - startCellX, \\
    offset_Y &= y - startCellY, \\
    offset_Z &= z - startCellZ.
\end{aligned}
\end{equation}
These offsets represent the position of the point within the convolution kernel.

\subsubsection{Position Mapping in the Convolved Space}

The resulting position in the convolved space is determined as:
\begin{equation}
\begin{aligned}
    o_x &= \frac{startCell_X}{s}, 
    o_y &= \frac{startCell_Y}{s}, 
    o_z &= \frac{startCell_Z}{s}.
\end{aligned}
\end{equation}

\subsubsection{Kernel Position Mapping}

The point's relative position within the kernel, considering its offset, is represented as:
\begin{equation}
    kernel\_offset = offset_X \times k^2 + offset_Y \times k + offset_Z,
\end{equation}
where $k$ denotes the kernel size.

\subsection{Submanifold Convolution}

In our approach, the computation process for sparse convolution can be concluded into three steps: create location table (LCT), create offset table (OFT) and compute result. The Submanifold convolution and sparse convolution has same steps in workflow, but there are little difference for inverse convolution because its special input structure and compute rules. In this part, we will introduce our methodology for Submanifold and others will be introduced in next part. Our solutions for Submanifold conlution are mainly follow the steps by works by Y. Yan, but we optimize the final computing process and data structure in computing:

\subsubsection{Create Location Table (LCT)}
In this step, we will scan the input indices to setup the relationship between a point take partcipate in computation and its location in inputs.

To linearize the 3D coordinates $(x, y, z)$ into a single index, the transformation is given by:
\begin{equation}
    \text{index} = x + y \times \text{max}_x + z \times \text{max}_x \times \text{max}_y
\end{equation}

The inverse transformation, which recovers the 3D coordinates from a given index, is described as:
\begin{align}
    z &= \left\lfloor \frac{\text{index}}{\text{max}_x \times \text{max}_y} \right\rfloor \\
    \text{index} &= \text{index} - z \times \text{max}_x \times \text{max}_y \\
    y &= \left\lfloor \frac{\text{index}}{\text{max}_x} \right\rfloor \\
    x &= \text{index} \mod \text{max}_x
\end{align}

In our work, The LCT can be stored in both CUDA Hash table \cite{alcantara2009real} or large CUDA array. For scenarios where the domain is relatively confined, allocating a three-dimensional CUDA array of dimensions X, Y, and Z is advantageous as its has better performance compared to CUDA hash table.

\begin{lstlisting}[caption=CUDA kernel for location table creation]
__global__ void createLCTKernel(int *locationTable, const int *n_values, const int *x_values, const int *y_values, const int *z_values, int N, int max_x, int max_y) {
    int idx = blockIdx.x * blockDim.x + threadIdx.x;
    if (idx < N) {
        int x = x_values[idx];
        int y = y_values[idx];
        int z = z_values[idx];
        locationTable[getIndex(x, y, z, max_x, max_y)] = idx;
    }
}
\end{lstlisting}

\subsubsection{Create Offset Table (OFT)} The offset table is a crucial element that guides convolution operations in sparse data environments, representing the relationship between spatial indices for efficient convolutions in sparse structures. The construction processes for SubM and Spconv operators differ significantly. A notable feature of the SubM convolution is its ability to preserve the congruity of input and output feature map dimensions, facilitating uninterrupted information flow. Identifying valid positions, central to the computation kernel, is vital. The rule table's creation depends on accurately defining spatial neighborhoods, influenced by the kernel size. Each voxel or point resides within a receptive field determined by the kernel dimensions, with the kernel's half-length defining its radial extent. Nested loops systematically explore the kernel's spatial range. At each point, the LCT is consulted to check for voxel activity. Active voxels are recorded in the rule table with their relative offsets, enhancing future data access efficiency.

\begin{figure}[!ht]
\centering
\begin{tikzpicture}[
    2d-arr/.style={matrix of nodes,
        row sep=-\pgflinewidth,
        column sep=-\pgflinewidth, 
        nodes={draw}}
  ]

\matrix (mtr) [2d-arr,  nodes={draw,
                      minimum width=1.8em,
                      minimum height=1.8em,
                      anchor=center,
                      inner sep=0pt,
                      outer sep=0pt,
                      font=\scriptsize},
               column sep=-\pgflinewidth,
               row sep=-\pgflinewidth] { 
$x_0$ & $y_0$ \\
 $\vdots$ & $\vdots$ \\ 
 |[fill=teal!30]|$31$ & |[fill=teal!30]|$31$  \\ 
 $\vdots$ & $\vdots$  \\ 
 |[fill=red!30]|$32$ & |[fill=red!30]|$32$ & \\ 
 $\vdots$ & $\vdots$  \\ 
 |[fill=gray!30]|$33$ & |[fill=gray!30]|$33$  \\
  $\vdots$ & $\vdots$  \\};

  \node[below=of mtr-7-2] {Input Indices};

  \matrix (K) [2d-arr, right=0.5em of mtr, nodes={draw,
                      minimum width=1.8em,
                      minimum height=1.8em,
                      anchor=center,
                      inner sep=0pt,
                      outer sep=0pt,
                      font=\scriptsize},
               column sep=-\pgflinewidth,
               row sep=-\pgflinewidth ] {
  0 & 0 & 0 & 0 & 0\\
  0 & |[fill=teal!30]|O & 0 & 0 &0\\
  0 & 0 & |[fill=red!30]|P & 0 & 0\\
  0 & 0 & 0 & |[fill=gray!30]|Q & 0 \\
  0 & 0 & 0 & 0 & 0\\
  };
  % \node[below=of K-3-2] {$\mathbf K$};

  \draw[dashed, teal] (mtr-3-2.north east) -- (K-2-2.north west);
  \draw[dashed, teal] (mtr-3-2.south east) -- (K-2-2.south west);

    \draw[dashed, teal] (mtr-5-2.north east) -- (K-3-3.north west);
  \draw[dashed, teal] (mtr-5-2.south east) -- (K-3-3.south west);

    \draw[dashed, teal] (mtr-7-2.north east) -- (K-4-4.north west);
  \draw[dashed, teal] (mtr-7-2.south east) -- (K-4-4.south west);

\matrix (ret) [matrix of nodes,
               nodes in empty cells,
               right=0.5em of K,
               nodes={draw,
                      minimum width=1.8em,
                      minimum height=1.8em,
                      anchor=center,
                      inner sep=0pt,
                      outer sep=0pt,
                      font=\scriptsize},
               column sep=-\pgflinewidth,
               row sep=-\pgflinewidth]
{
$o_0^0$ & $\dots$ & $o_0^4$ & $\dots$ & $o_0^8$\\
$\vdots$ & $\vdots$ & $\vdots$ & $\vdots$ & $\vdots$\\ 
|[fill=teal!30]| -1 & |[fill=teal!30]| $\dots$ & |[fill=teal!30]| $I_O$ & |[fill=teal!30]| $\dots$ & |[fill=teal!30]| $I_P$\\ 
$\vdots$ & $\vdots$ & $\vdots$ & $\vdots$ & $\vdots$\\ 
|[fill=red!30]| $I_O$ & |[fill=red!30]| $\dots$ & |[fill=red!30]| $I_P$ & |[fill=red!30]| $\dots$ & |[fill=red!30]| $I_Q$\\ 
$\vdots$ & $\vdots$ & $\vdots$ & $\vdots$ & $\vdots$\\ 
|[fill=gray!30]| $I_P$ & |[fill=gray!30]| $\dots$ & |[fill=gray!30]| $I_Q$ & |[fill=gray!30]| $\dots$ & |[fill=gray!30]| -1\\ 
$\vdots$ & $\vdots$ & $\vdots$ & $\vdots$ & $\vdots$\\
};
 
  \node[below=of ret-7-3] {Result OFT};

  \draw[dashed, teal] (ret-3-1.north west) -- (K-2-2.north east);
  \draw[dashed, teal] (ret-3-1.south west) -- (K-2-2.south east);

    \draw[dashed, teal] (ret-5-1.north west) -- (K-3-3.north east);
  \draw[dashed, teal] (ret-5-1.south west) -- (K-3-3.south east);

    \draw[dashed, teal] (ret-7-1.north west) -- (K-4-4.north east);
  \draw[dashed, teal] (ret-7-1.south west) -- (K-4-4.south east);

  % \draw[dashed, teal] (mtr-1-1.south west) -- (K-1-1.north east);
  % \draw[dashed, teal] (mtr-1-5.south east) -- (K-5-1.south east);

  % \draw[dashed, blue!80!black] (mtr-1-5.south west) -- (ret-1-1.north west);
  % \draw[dashed, blue!80!black] (mtr-1-1.south east) -- (ret-1-1.south west);

\end{tikzpicture}
\caption{Visualization of the rule construction process for a submanifold convolution. The 'Input Indices' matrix represents the non-zero entries of the input feature map. The process involves scanning these non-zero entries to identify their valid neighboring non-zero entries. The 'Result OFT' (Offset Table) captures the spatial relations of these entries by setting the corresponding locations within the table, thereby encoding the connectivity patterns necessary for the convolution operation.}
\label{fig_sim}
\end{figure}

The convolution process for Spconv presents a more intricate challenge when compared to SubM, particularly in the construction and management of the Offset Table (OFT). This complexity arises due to the inherent nature of Spconv, where the shapes of input and output feature maps aren't identical, introducing a variable dimensionality into the convolution operations. An extra step are added in steps of building LOT, which will count the number of down-sample points and get the shape of output tensor. In the output space, one point may contributed by many points in origin input space, it raise a predominant challenge that emerges during the Spconv process is managing and mitigating CUDA race conditions. Specifically, the nature of parallelized GPU operations, where multiple threads might attempt to access or modify shared memory locations simultaneously, introduces potential conflicts

To handle this, we use The primary goal of the spconv\_createLookupKernel is to craft a mapping scheme – a "Rule-Mapping Book" that correlates the input and output spaces, incorporating the variances introduced by parameters like kernel size and stride. Once this mapping is established, it's imperative to synchronize operations to ensure that the Rule-Mapping Book is fully populated before subsequent operations, thereby ensuring data integrity and consistency.

This challenge is exemplified in the setUniquePoints kernel. The kernel's primary objective is to uniquely identify the spatial indices in the feature map and populate the d\_idxList and d\_offsetTable accordingly. Given that multiple threads might simultaneously operate on overlapping or neighboring spatial regions, there's a risk of threads overwriting each other’s data or reading inconsistent data states, leading to inaccuracies in the resultant OFT.

\begin{algorithm}
\caption{Normal Convolution Computation}
\begin{algorithmic}[1]
\REQUIRE \textbf{Input:} $inF, rT, w, outF, N, inC, outC, kS$ \COMMENT{F represents Features, C represents Channels, rT is RulesTable, kS is kernel size}
\ENSURE \textbf{Output:} $outF$ after convolution

\STATE $kV \gets kS^3$
\FOR{$i = 0$ \TO $N-1$}
    \FOR{$o = 0$ \TO $outC-1$}
        \STATE $s \gets 0$
        \FOR{$k = 0$ \TO $kV-1$}
            \STATE $nIdx \gets rT[i \cdot kV + k]$
            \IF{$nIdx \neq -1$}
                \FOR{$c = 0$ \TO $inC-1$}
                    \STATE $wIdx \gets o \cdot kV \cdot inC + k \cdot inC + c$
                    \STATE $s \gets s + inF[nIdx \cdot inC + c] \cdot w[wIdx]$
                \ENDFOR
            \ENDIF
        \ENDFOR
        \STATE $outF[i \cdot outC + o] \gets s$
    \ENDFOR
\ENDFOR
\end{algorithmic}
\end{algorithm}

\subsubsection{Compute Result} This step will execute computation process for output using OFT and weights. The original method, as described in Algorithm 2, adopts a nested loop approach to compute the convolution values. While conceptually straightforward, this technique exhibits inefficiencies when implemented on a GPU architecture. One of the main limitations is the frequent access to the global memory, especially for retrieving weight values. Since global memory access is notably slower than shared or local memory access on CUDA devices, this becomes a performance bottleneck. The nested nature of the loops amplifies this inefficiency, leading to suboptimal performance, especially for larger volumes of data.

\begin{figure}[h]

\begin{minipage}{0.24\textwidth}
\centering

\begin{tikzpicture}[
    2d-arr/.style={matrix of nodes,
        row sep=-\pgflinewidth,
        column sep=-\pgflinewidth, 
        nodes={draw}}
  ]

\matrix (mtr) [2d-arr,nodes={draw,
                      minimum width=1.8em,
                      minimum height=1.8em,
                      anchor=center,
                      inner sep=0pt,
                      outer sep=0pt,
                      font=\scriptsize},
               column sep=-\pgflinewidth,
               row sep=-\pgflinewidth] { 
 |[fill=pink!30]|$o_1^1$ & |[fill=pink!30]|$o_1^2$ & |[fill=pink!30]|$o_1^3$ & |[fill=pink!30]|$\cdots$ & |[fill=pink!30]|$o_1^c$ \\ 
 $o_2^1$ & $o_2^2$ & $o_2^3$ & $\cdots$ & $o_2^c$ \\ 
 $o_3^1$ & $o_3^2$ & $o_3^3$ & $\cdots$ & $o_3^c$ \\ 
 $o_4^1$ & $o_4^2$ & $o_4^3$ & $\cdots$ & $o_4^c$ \\ 
 $o_5^1$ & $o_5^2$ & $o_5^3$ & $\cdots$ & $o_5^c$ \\ 
 $\vdots$ & $\vdots$ & $\vdots$ & $\vdots$ & $\vdots$ \\ 
 $o_n^1$ & $o_n^2$ & $o_n^3$ & $\cdots$ & $o_n^c$ \\  };

  \node[below=of mtr-6-3] {Offset Table (Features)};

  % \node[below=of mtr--3] {$\mathbf I$};

  \node[left=0.2em of mtr] (str) {$*$};

  \matrix (K) [2d-arr, left=0.1em of str, nodes={draw, fill=green!30, minimum width=2em, minimum height=2em}] {
    $w_1$\\
    $w_2$\\
    $w_3$\\
    $\vdots$\\
    $w_c$\\
  };
  \node[below=of K-4-1] {Weights};

  \node[right=0.2em of mtr] (eq) {$=$};

  \matrix (ret) [2d-arr, right=0.1em of eq, nodes={draw,
                      minimum width=1.8em,
                      minimum height=1.8em,
                      anchor=center,
                      inner sep=0pt,
                      outer sep=0pt,
                      font=\scriptsize},
               column sep=-\pgflinewidth,
               row sep=-\pgflinewidth] {
    |[fill=brown!30]|$r_1$\\
    $r_2$\\
    $r_3$\\
    $r_4$\\
    $r_5$\\
    $\vdots$\\
    $r_n$\\
  };
  \node[below=of ret-6-1] {Results};

  % \draw[dashed, teal] (mtr-1-1.south west) -- (K-1-1.north east);
  % \draw[dashed, teal] (mtr-1-5.south east) -- (K-5-1.south east);

  % \draw[dashed, blue!80!black] (mtr-1-5.south west) -- (ret-1-1.north west);
  % \draw[dashed, blue!80!black] (mtr-1-1.south east) -- (ret-1-1.south west);

\end{tikzpicture}

\end{minipage}

\vspace{0.5cm}

\centering
\begin{minipage}{0.24\textwidth}
    \centering

\begin{tikzpicture}[
    2d-arr/.style={matrix of nodes,
        row sep=-\pgflinewidth,
        column sep=-\pgflinewidth, 
        nodes={draw}}
  ]

\matrix (mtr) [2d-arr,nodes={draw,
                      minimum width=1.8em,
                      minimum height=1.8em,
                      anchor=center,
                      inner sep=0pt,
                      outer sep=0pt,
                      font=\scriptsize},
               column sep=-\pgflinewidth,
               row sep=-\pgflinewidth] { 
 |[fill=blue!30]|$r_1^1$ & |[fill=yellow!30]|$r_1^2$ & |[fill=brown!30]|$r_1^3$ & |[fill=lime!30]|$\cdots$ & |[fill=teal!30]|$r_1^c$ \\ 
 |[fill=blue!30]|$r_2^1$ & |[fill=yellow!30]|$r_2^2$ & |[fill=brown!30]|$r_2^3$ & |[fill=lime!30]|$\cdots$ & |[fill=teal!30]|$r_2^c$ \\ 
 |[fill=blue!30]|$r_3^1$ & |[fill=yellow!30]|$r_3^2$ & |[fill=brown!30]|$r_3^3$ & |[fill=lime!30]|$\cdots$ & |[fill=teal!30]|$r_3^c$ \\ 
 |[fill=blue!30]|$r_4^1$ & |[fill=yellow!30]|$r_4^2$ & |[fill=brown!30]|$r_4^3$ & |[fill=lime!30]|$\cdots$ & |[fill=teal!30]|$r_4^c$ \\ 
 |[fill=blue!30]|$r_5^1$ & |[fill=yellow!30]|$r_5^2$ & |[fill=brown!30]|$r_5^3$ & |[fill=lime!30]|$\cdots$ & |[fill=teal!30]|$r_5^c$ \\ 
 |[fill=blue!30]|$\vdots$ & |[fill=yellow!30]|$\vdots$ & |[fill=brown!30]|$\vdots$ & |[fill=lime!30]|$\vdots$ & |[fill=teal!30]|$\vdots$ \\ 
 |[fill=blue!30]|$r_n^1$ & |[fill=yellow!30]|$r_n^2$ & |[fill=brown!30]|$r_n^3$ & |[fill=lime!30]|$\cdots$ & |[fill=teal!30]|$r_n^c$ \\  
};

% x-axis
\draw[thick,->] ([yshift=0em]mtr.north west) -- ([yshift=0em]mtr.north east) node[right] {$grid.y$};
% y-axis
\draw[thick,->] ([xshift=-0em]mtr.north west) -- ([xshift=-0em]mtr.south west) node[below] {$grid.x$};

\end{tikzpicture}

    \end{minipage}

\caption{The upper portion illustrates the computation of convolution from a given set of weights and a offset table. The offset table, representing features, directs the mapping of weights to corresponding feature values by indexed rules, facilitating the convolution process. The lower portion depicts the CUDA grid configuration and the allocation of weights in shared memory. Each output channel is assigned to a distinct CUDA grid dimension, with its weights stored in shared memory to expedite the convolution operation.}
\end{figure}

The optimized method offers a more efficient approach. By assigning the output channel, $oc$, directly to the block's y-dimension, the nested loop for output channels is eliminated. Furthermore, the most significant enhancement is the introduction of shared memory, which caches the `weights` for faster access. This approach dramatically reduces the frequency of slow global memory accesses. Loading the `weights` into shared memory ensures that all threads within a block can access the values with reduced latency, speeding up the convolution computation considerably.

\begin{lstlisting}
__global__ void computeConv(const float *input, 
    const int *rules, const float *w, float *output, 
    int num, int in_ch, int out_ch, int k_size) {
    int idx = blockIdx.x * blockDim.x + threadIdx.x;
    int oc = blockIdx.y;
    int k_vol = k_size * k_size * k_size;
    extern __shared__ float sW[];
    for (int k = 0; k < k_vol; k++) {
        for (int ic = 0; ic < in_ch; ic++) {
            int w_idx = oc * k_vol * in_ch + k * in_ch + ic;
            sW[k * in_ch + ic] = w[w_idx];
        }
    }
    __syncthreads();

    if (idx < num) {
        float sum = 0;
        for (int k = 0; k < k_vol; k++) {
            int nIdx = rules[idx * k_vol + k];
            if (nIdx != -1) {
                for (int ic = 0; ic < in_ch; ic++) {
                    sum += input[nIdx * in_ch + ic] * sW[k * in_ch + ic];
                }
            }
        }
        output[idx * out_ch + oc] = sum;
    }
}

\end{lstlisting}

\subsection{Solution for Normal Sparse Convolution}
As there are might many position map to a same point after down sample process, there are two crucial aspects we need to consider in design sparse convolution operator: (1) How to get the amount of elements after down sample during the parallel computing process, which determines the output size and shape of offset table. (2) Build suitable rules table for them. To handle this, we used a counter to record the output size by combination of CUDA atomic opeation and a temp table to record a position after downsample exists, and its location in output list. In addition, for each location in origin input, we used a new-design two-stage offset table. 

In our counter design, it contains two steps: (1) The first step involves an atomic exchange operation. Each thread, working on a distinct set of input coordinates, computes an index corresponding to a potential output position in the downsampled space. To ascertain whether this position has already been flagged by another thread, the algorithm performs an atomic exchange operation on a shared status table. This table is pre-initialized with a specific value (e.g., -1) to indicate unoccupied positions. The atomic exchange operation attempts to replace the value at the computed index with a new value, signifying that the position is now occupied. Crucially, this operation returns the original value at the index, allowing the thread to determine whether the position was previously unoccupied. If the returned value from the atomic exchange is the pre-initialized one (indicating an unoccupied state), the thread has successfully identified a unique output position. (2) In the second step, the thread atomically increments a global counter. This counter keeps track of the total number of unique output positions identified across all threads. The atomic increment ensures that even when multiple threads simultaneously identify unique positions, the count remains accurate and consistent.

In the step (1) at above process, we create a table contains two columns. The first column records index in origin input second column records its offset in kernel. After that, according to above table and size of output, then we can set offset table with same schema in above part used for convolution computing. Finally, we apply above convolution solution to get output.

\subsection{Design for Inverse Convolution}

\begin{figure}[ht]
\includegraphics[width=8cm]{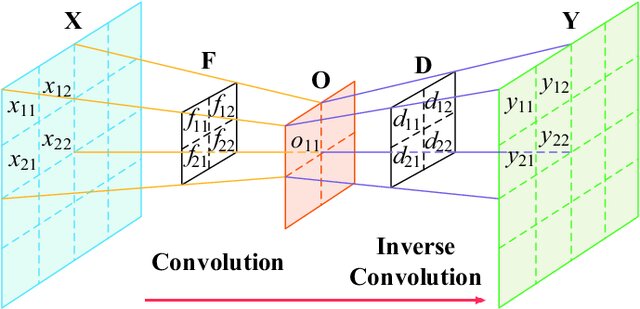}
\centering
\caption{Schematic diagram of convolution and inverse convolution process, source: \cite{articledeconv}. The inverse convolution essentially redistributes the values from O across a larger matrix using the filter D, aiming to approximate the original matrix X's structure and values in Y.}
\label{fig:inv_conv}
\end{figure}

The process of deconvolution, which is the mathematical inverse of convolution, plays a crucial role in the technique of up-sampling. Up-sampling is a method used to enhance the data rate or resolution. By employing the technique of deconvolution, it becomes possible to recreate a dataset with higher resolution by reversing the convolutional procedure that was applied to its downsampled version. Figure \ref{fig:inv_conv} shows process of inverse convolution.

 In contrast to conventional dictionary-based methods, our approach dynamically generates associations between input and output data elements in the context of sparse convolution, in real-time. This is achieved through the utilization of two distinct sets of indices: \emph{indices1} and \emph{indices2} which represent the feature indices before and after convolution respectively. We make the assumption that the size of \emph{indices1} is greater than \emph{indices2}, and by leveraging this aspect, we deduce that \emph{indices2} can be inferred from \emph{indices1} coupled with the convolution parameters such as kernel size and stride.

Expanding on this concept, we detail a method for reversing the convolution operation. The initial phase involves constructing a computation rules table, analogous in size to the pre-convolution state. In the context of inverse convolution, the computation focuses on singular weight and feature pairs. The rules table is organized such that the first column catalogues the feature values, while the second column details the weight values at specific kernel positions. Consequently, the inverse convolution computation simplifies to the product of these two values. Subsequent stages require iterative scanning through all coordinates $(x_i,y_i,z_i)$ belonging to \emph{indices1}. We then locate the corresponding location within \emph{indices2} that share a relation with $(x_i,y_i,z_i)$ and its offset in the convolution kernel location so that we can put feature value for  $(x_i,y_i,z_i)$ and its weights in specific kernel offset in a newly constructed feature map. Final stages involve applying a new convolution kernel on this feature map resulting in a higher resolution data representation.

To summarize, our proposed methodology is laid out in three distinct stages that involve generation of location table for \emph{indices2}, assignment of input feature values to appropriate locations in a temporary feature array and the implementation of the inverse convolution operation. This approach thrives on the careful balancing of computational resources and improves data accuracy by means of advanced kernel weighting during the convolution operation.

\section{Conclusion}

We mainly contribute two improvement in the implement of sparse convolution on mobile embedded systems: (1) Optimize convolution computation process utlizing the advantage of CUDA. (2) A easy and efficient approch to implement inverse sparse convolution. This work effectively presents an innovative implementation of sparse convolution operators using CUDA, emphasizing maximized parallelism and efficient data load optimization. It introduces a novel approach to handling tensor structures within CUDA's framework, aligning with PyTorch's usability while leveraging CUDA's performance capabilities. The methodology, encompassing the creation of location table and offset table followed by computation, optimizes sparse convolution processes, particularly addressing inefficiencies in traditional nested loop approaches. By incorporating shared memory for weight caching and optimizing thread and block utilization, the thesis demonstrates significant advancements in the efficiency and speed of sparse convolution operations on GPU architectures, offering substantial contributions to the field of GPU-accelerated computing.

% if have a single appendix:
%\appendix[Proof of the Zonklar Equations]
% or
%\appendix  % for no appendix heading
% do not use \section anymore after \appendix, only \section*
% is possibly needed

% use appendices with more than one appendix
% then use \section to start each appendix
% you must declare a \section before using any
% \subsection or using \label (\appendices by itself
% starts a section numbered zero.)
%

% \appendices
% \section{Proof of the First Zonklar Equation}
% Appendix one text goes here.

% % you can choose not to have a title for an appendix
% % if you want by leaving the argument blank
% \section{}
% Appendix two text goes here.

% % use section* for acknowledgment
% \section*{Acknowledgment}

% Can use something like this to put references on a page
% by themselves when using endfloat and the captionsoff option.
\ifCLASSOPTIONcaptionsoff
  \newpage
\fi

% trigger a \newpage just before the given reference
% number - used to balance the columns on the last page
% adjust value as needed - may need to be readjusted if
% the document is modified later
%\IEEEtriggeratref{8}
% The "triggered" command can be changed if desired:
%\IEEEtriggercmd{\enlargethispage{-5in}}

% references section

% can use a bibliography generated by BibTeX as a .bbl file
% BibTeX documentation can be easily obtained at:
% http://mirror.ctan.org/biblio/bibtex/contrib/doc/
% The IEEEtran BibTeX style support page is at:
% http://www.michaelshell.org/tex/ieeetran/bibtex/
%\bibliographystyle{IEEEtran}
% argument is your BibTeX string definitions and bibliography database(s)
%\bibliography{IEEEabrv,../bib/paper}
%
% <OR> manually copy in the resultant .bbl file
% set second argument of \begin to the number of references
% (used to reserve space for the reference number labels box)

\bibliographystyle{IEEEtran}
\bibliography{IEEEabrv, bibtex/bib/IEEEexample}
% biography section
% 
% If you have an EPS/PDF photo (graphicx package needed) extra braces are
% needed around the contents of the optional argument to biography to prevent
% the LaTeX parser from getting confused when it sees the complicated
% \includegraphics command within an optional argument. (You could create
% your own custom macro containing the \includegraphics command to make things
% simpler here.)
%\begin{IEEEbiography}[{\includegraphics[width=1in,height=1.25in,clip,keepaspectratio]{mshell}}]{Michael Shell}
% or if you just want to reserve a space for a photo:

% \begin{IEEEbiography}{Michael Shell}
% Biography text here.
% \end{IEEEbiography}

% if you will not have a photo at all:
% \begin{IEEEbiographynophoto}{John Doe}
% Biography text here.
% \end{IEEEbiographynophoto}

% % insert where needed to balance the two columns on the last page with
% % biographies
% %\newpage

% \begin{IEEEbiographynophoto}{Jane Doe}
% Biography text here.
% \end{IEEEbiographynophoto}

% You can push biographies down or up by placing
% a \vfill before or after them. The appropriate
% use of \vfill depends on what kind of text is
% on the last page and whether or not the columns
% are being equalized.

%\vfill

% Can be used to pull up biographies so that the bottom of the last one
% is flush with the other column.
%\enlargethispage{-5in}

% that's all folks
\end{document}